\documentclass[letterpaper, 10 pt, conference]{ieeeconf}  

\IEEEoverridecommandlockouts                              

\usepackage{epsfig} 
\usepackage{times} 
\usepackage{amsmath} 

\usepackage{amssymb}  
\usepackage{stfloats}
\usepackage{graphicx}
\usepackage{float}
\usepackage{makecell}
\usepackage{multirow}
\usepackage{booktabs} 
\usepackage{cite}     
\usepackage[ruled,linesnumbered]{algorithm2e}[ruled,linesnumbered]
\usepackage{booktabs}
\usepackage[table]{xcolor} 
\definecolor{green}{rgb}{0.96, 0.76, 0.75}
\definecolor{yellow}{rgb}{0.98, 0.94, 0.31}
\usepackage{threeparttable}
\usepackage[table]{xcolor}

\title{\LARGE \bfseries
Decentralized Multi-Robot Relative Navigation in Unknown, Structurally Constrained Environments under Limited Communication
}

\begin{document}
\author{ Zihao Mao, Yunheng Wang, Yunting Ji, Yi Yang, Wenjie Song\textsuperscript{*}  
\thanks{This work was partly supported by Program for National Natural Science
Foundation of China (Grant No. 62373052), Beijing Natural Science Foundation (Grant No. 4252051), and in part by the National Key Laboratory of
Science and Technology on Space Born Intelligent Information Processing
TJ-01-22-09.}
\thanks{The authors are with the School of Automation, Beijing Institute of
Technology, Beijing 100081, China, (Corresponding author: Wenjie Song,
email: songwj@bit.edu.cn).}%
}

\maketitle
\thispagestyle{empty}
\pagestyle{empty}

\begin{abstract}

Multi-robot navigation in unknown, structurally constrained, and GPS-denied environments presents a fundamental trade-off between global strategic foresight and local tactical agility, particularly under limited communication. Centralized methods achieve global optimality but suffer from high communication overhead, while distributed methods are efficient but lack the broader awareness to avoid deadlocks and topological traps.
To address this, we propose a fully decentralized, hierarchical relative navigation framework that achieves both strategic foresight and tactical agility without a unified coordinate system. 
At the strategic layer, robots build and exchange lightweight topological maps upon opportunistic encounters. This process fosters an emergent global awareness, enabling the planning of efficient, trap-avoiding routes at an abstract level. This high-level plan then inspires the tactical layer, which operates on local metric information. Here, a sampling-based ‘escape point’ strategy resolves dense spatio-temporal conflicts by generating dynamically feasible trajectories in real time, concurrently satisfying tight environmental and kinodynamic constraints.
Extensive simulations and real-world experiments demonstrate that our system significantly outperforms in success rate and efficiency, especially in communication-limited environments with complex topological structures.

\end{abstract}

\section{INTRODUCTION}

Effective navigation and coordination in multi-robot systems remain a persistent challenge, particularly in unknown, structurally-complex, and GPS-denied environments \cite{app1,app2}. In such scenarios, limited communication severely restricts the information flow essential for team collaboration. This information scarcity precludes the formulation of efficient global navigation strategies, often yielding paths that fail to consider the broader environmental topology. Consequently, robots are frequently steered into constricted areas, leading to dense local conflicts and deadlocks that demand real-time resolution under strict environmental and kinodynamic constraints. 
Inefficient global planning is prone to steering the team in these locally conflicted, hard-to-resolve situations. This coupling of myopic global planning with complex local interactions creates a failure cascade, making the system highly susceptible to gridlock and mission failure.

Existing approaches, whether centralized or distributed, have struggled to effectively address this challenge because they fail to reconcile the trade-off of balancing global, long-term strategic planning with local, real-time conflict resolution.
On one hand, reactive distributed methods such as Reciprocal Velocity Obstacles \cite{rvo,rvo1,rvo2, rvo3,rvo4} and Distributed Model Predictive Control \cite{dmpc,dmpc1}, are computationally efficient and excel at local, real-time collision avoidance. However, their reliance on purely local metric data renders them fundamentally short-sighted. Without a shared understanding of the global environment, these robots are prone to strategically inefficient paths, susceptible to large-scale topological traps (e.g., large U-shaped obstacles), and often fail to resolve complex deadlock scenarios.
On the other hand, methods aiming for global coordination and optimal, deadlock-free paths typically build a shared, unified coordinate system by exchanging dense metric maps, such as occupancy grids \cite{cbs}. While this approach provides the global awareness needed to avoid large-scale obstacles, its construction of a globally consistent metric reference frame relies heavily on continuous data streams. This makes the system highly vulnerable to network disruptions. Additionally, it exhibits poor flexibility in local obstacle avoidance, low efficiency, and poor scalability, as each planning iteration requires coordination among all robots. Collectively, these challenges severely restrict the practicality of such methods in real-world deployments.

\begin{figure}[tbp]
    \centering
    \includegraphics[scale=0.16]{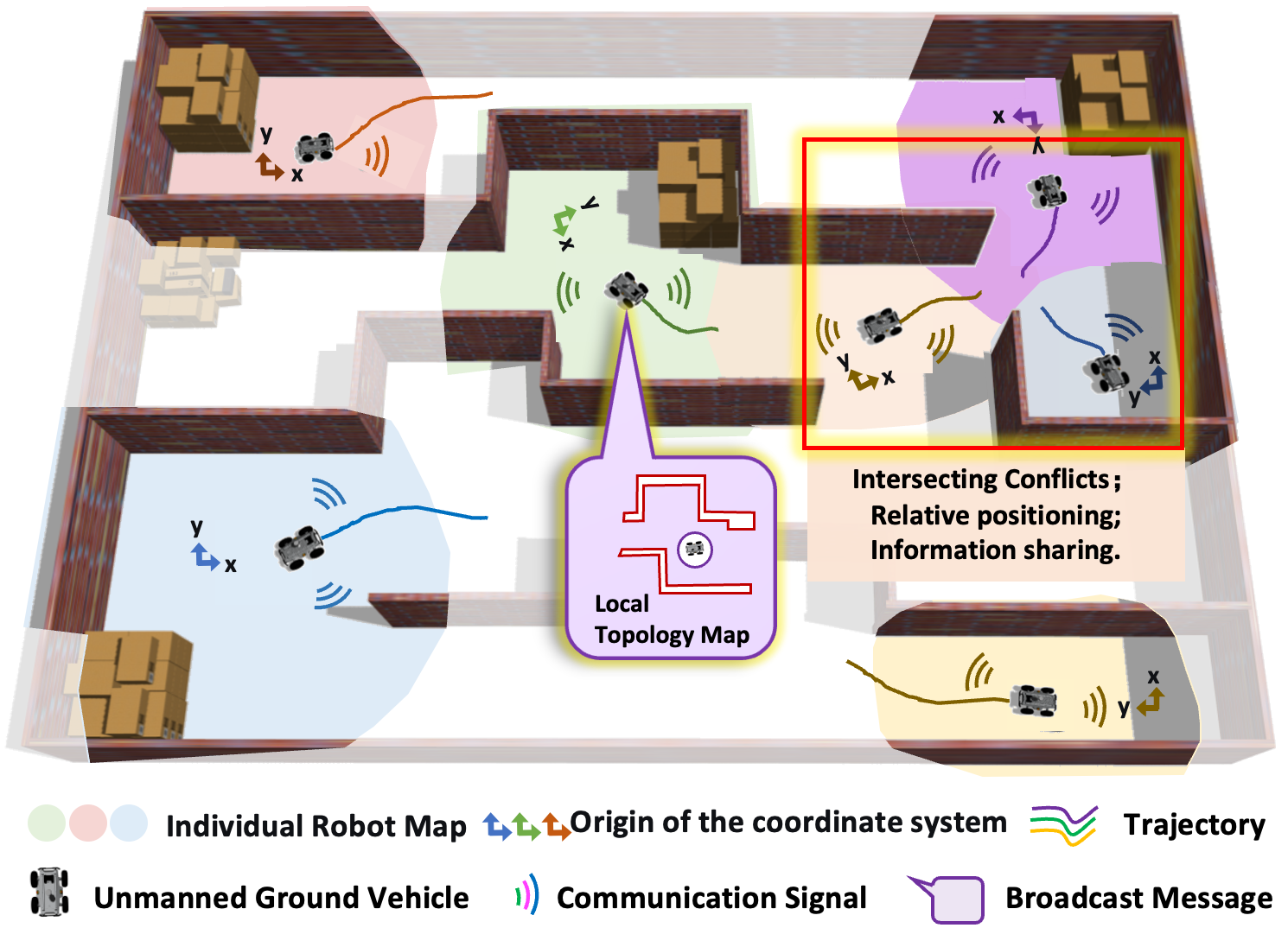}
    \caption{Our hierarchical framework enables a team of robots to navigate efficiently in topologically complex environments with limited communication. Global awareness emerges from sharing abstract topological maps, which guides a local metric-level planner that ensures agile and safe collision avoidance.}
    \label{fig1}
    \vspace{-1em}
\end{figure}

To overcome these limitations, our core insight is that robust, global-scale coordination can emerge from sparse, local, and pairwise interactions, thus abandoning the need for a costly and fragile unified coordinate system. 
We introduce a hierarchical framework built on a relative navigation paradigm, that achieves the strategic foresight of centralized systems with the agility and communication efficiency of distributed methods. As shown in the schematic diagram in Fig.~\ref{fig1}, at its higher, strategic topological layer, each robot independently explores and maintains its own topological map, represented as a visibility graph. The key to achieving global awareness lies in pairwise information exchange: upon encounter, teammates perform onboard sensor-based relative localization and share their lightweight topological maps. Each robot then integrates the received topology into its own map, creating a richer ‘roadmap’ of the world extending far beyond its direct sensing horizon. This decentralized process allows the team to enhance navigation efficiency and avoid large-scale topological traps.
At the lower, tactical metric layer, this high-level topological path inspires a real-time motion planner that operates on local metric sensor data. This layer generates smooth, dynamically feasible trajectories that adhere to the global strategy while ensuring immediate safety. To resolve dense spatio-temporal conflicts in cluttered spaces, we introduce a rapid conflict-resolution strategy. This method performs online sampling to generate kinodynamically feasible ‘escape points’ in unoccupied regions of the joint state-space. These points act as immediate sub-goals, allowing robots to proactively and safely maneuver around one another, ensuring fluid navigation.
The main contributions of this work are as follows.
\begin{itemize}
    \item \textbf{A hierarchical relative navigation framework} that synergizes lightweight topological sharing for long-horizon strategic planning and local metric-based planning for real-time collision avoidance, eliminating the need for a shared global coordinate system.
    \item \textbf{A decentralized map-sharing method} where global navigational awareness emerges from local, pairwise lightweight topological map fusions, enhancing team efficiency and preventing entrapment in complex topological traps under sparse communication.
    \item \textbf{A rapid, kinodynamically-aware conflict-resolution strategy} that generates ‘escape points’ online, enabling robots to find safe and efficient trajectories around local conflicts under tight environmental and dynamic constraints at the metric level.
\end{itemize}

We validated our framework through simulations and real-world experiments in challenging and structurally complex environments, demonstrating superior performance in success rate and navigation efficiency compared to state-of-the-art methods.

\section{RELATED WORK}
\subsection{multi robot navigation}
Multi-robot navigation in unknown environments requires a delicate balance between collision avoidance, deadlock resolution, and strategic planning. 
Distributed reactive methods, exemplified by Reciprocal Velocity Obstacles (RVO/ORCA) \cite{rvo,rvo1,rvo3,rvo2,rvo4, rvo5} and Distributed Model Predictive Control (DMPC) \cite{dmpc,dmpc1}, are computationally lightweight and excel at generating smooth, collision-free local trajectories in real-time. However, their reliance on purely local information renders them fundamentally short-sighted. Lacking communication or a shared global understanding, they are highly susceptible to deadlocks in dense scenarios and can guide robots into large-scale topological traps from which recovery is difficult \cite{rvo4}.
In contrast, centralized planners like, such as \cite{cbs,cen1,cen2,cen3,cen4, cen5}, which leverage a global view to compute certifiably complete and optimal paths, effectively resolving complex multi-robot conflicts. This global optimality, however, is achieved at a significant cost. Their dependence on a unified coordinate system and a central processor results in prohibitive communication overhead and computational complexity, rendering them unscalable and brittle in communication-constrained, dynamic environments. While hybrid methods that combine Conflict-Based Search \cite{cbs} with local planners exist \cite{cen5}, they still inherit the fundamental bottleneck of centralization.
Our work combines the advantages of both categories in a distributed form. By synergizing a lightweight global topological layer with a local reactive layer, we achieve the long-range foresight of centralized planners via lightweight topological exchange without their communication overhead, while retaining the agility and scalability of reactive methods.

\subsection{World Representation}

Classical multi-robot coordination often assumes a known global map and precise localization, which is seldom feasible in our target scenarios. More recent methods attempt to construct this global map online, but this typically requires sharing dense metric data like occupancy grids. This process is vulnerable to high bandwidth demands and can suffer catastrophic failures from accumulated odometry drift.
An alternative is seen in Reinforcement Learning, where robots operate on relative observations \cite{rl1, rl2, rl3, rl4}. Architectures using centralized training with distributed execution (CTDE) \cite{rl4} share conceptual similarities with our paradigm by using relative coordinates. However, learning-based methods often suffer from poor generalization to novel scenarios and can be unreliable in dense, safety-critical conflict situations. Our framework circumvents these issues by representing the world with a lightweight topological structure, which is more scalable and robust for strategic planning.

\subsection{Local Conflict Resolution}
Even with a global plan, resolving immediate, dense conflicts remains a challenge. Simple heuristics like the right-hand rule \cite{con_res1,con_res2,con_res3} or fixed priorities are insufficient, as they often fail or cause cascading conflicts in structured environments like narrow corridors. In contrast to these rigid rules, our local planner employs a more sophisticated, sampling-based escape strategy. Crucially, this strategy is guided by the global topological route, ensuring that tactical avoidance maneuvers remain consistent with the long-term strategic goal. This synergy prevents the planner from making locally safe but globally detrimental decisions, a key limitation of purely reactive approaches.

\section{Preliminaries}
\subsection{Task Definition}

We consider a team of $N$ Unmanned Ground Vehicles (UGVs), $\mathcal{A} = \{1, \dots, N\}$, operating in an a priori unknown workspace $\mathcal{W} \subset \mathbb{R}^2$ containing obstacles $\mathcal{O}$ and complex topological features. Each UGV $i \in \mathcal{A}$ is tasked with navigating from a start state $x_i^{\text{start}}$ to a goal position $p_i^{\text{goal}}$.
The team operates under fully decentralized constraints: (i) robots lack access to a global coordinate system or GPS; (ii) they rely solely on onboard sensing for local perception; and (iii) communication is limited to opportunistic, peer-to-peer exchanges within a finite range. The objective is to generate a set of safe, efficient, and dynamically feasible trajectories, that guide each robot to its goal without colliding with obstacles or other robots. This problem requires balancing long-horizon strategic pathfinding to avoid large-scale topological traps with real-time tactical maneuvering to resolve local conflicts.

\begin{figure*}[t]
    \centering
    \includegraphics[scale=0.18]{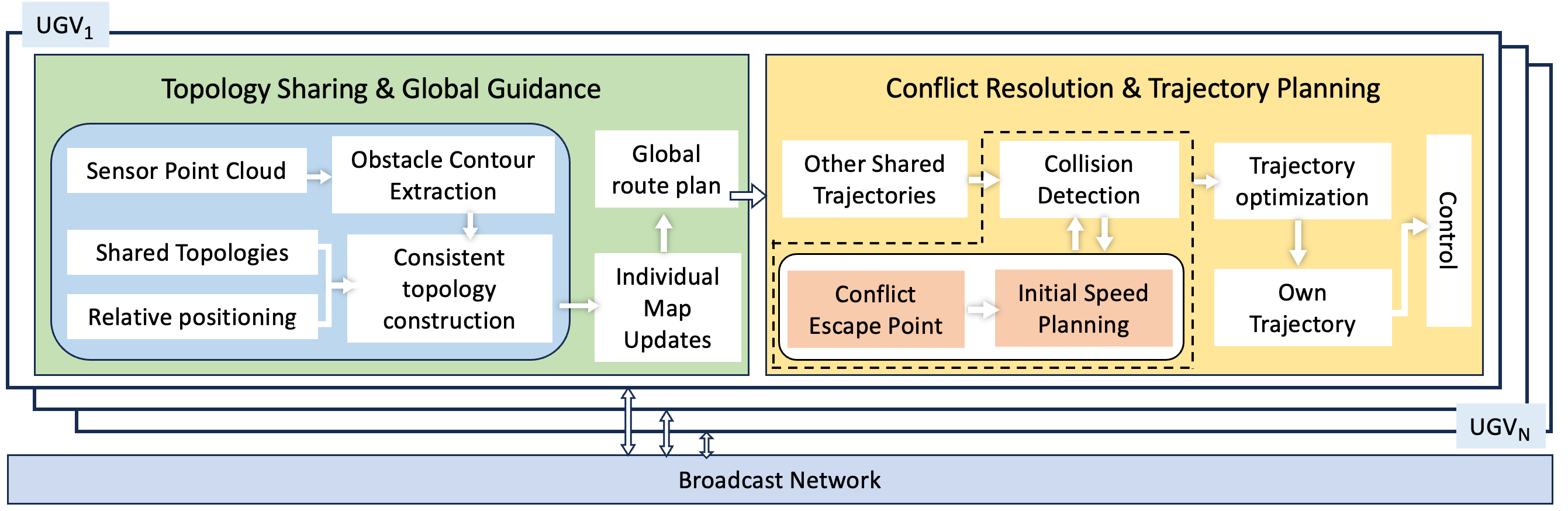}
    \caption{Overview of our proposed decentralized, hierarchical navigation framework. Each robot runs an independent planner composed of a high-level Topology Sharing \& Global Guidance (TSGG) layer for strategic planning and a low-level Conflict Resolution \& Trajectory Planning (CRTP) layer for tactical execution.}
    \label{system framework}
    \vspace{-1em}
\end{figure*}

\subsection{System Framework}
To address the challenges above, we propose the decentralized, hierarchical navigation framework illustrated in Fig.~\ref{system framework}. It is designed to enable robust and efficient cooperation in large-scale, unknown environments with complex features such as narrow corridors and dead ends. At its core, each UGV independently operates a planner composed of two symbiotic modules: a high-level \textbf{Topology Sharing 
\& Global Guidance (TSGG)} module and a low-level \textbf{Conflict Resolution \& Trajectory Planning (CRTP)} module.

The TSGG module is responsible for establishing long-horizon strategic awareness. Operating at a low frequency, it constructs a lightweight, topological map of the environment. Through opportunistic, pairwise exchange of these sparse representations, the team collectively develops an understanding of the environment's connectivity. This shared knowledge enables the generation of strategically sound, trap-avoiding global routes without a unified metric frame.

This high-level guidance inspires the CRTP layer, which operates at a high frequency to manage immediate, dynamic interactions. Using the strategic route from the TSGG as a reference, the CRTP module generates a smooth, dynamically feasible, and locally collision-free trajectory. To coordinate with nearby robots, it shares state information and employs a rapid conflict resolution strategy to resolve any impending spatio-temporal conflicts in real time.

The efficacy of this framework stems from its strategic decoupling of the planning problem. The low-frequency, low-bandwidth sharing of topological maps addresses the global navigation challenge (e.g., avoiding dead ends and major detours), while the high-frequency local motion planner, operating on real-time sensor data, solves the immediate coordination problem (e.g., avoiding collisions). This hierarchical design enables the team to achieve both far-sighted global intelligence and real-time local reactivity, all within a communication-efficient, decentralized paradigm.

\section{Method}
\begin{figure}[t]
    \centering
    \includegraphics[scale=0.42]{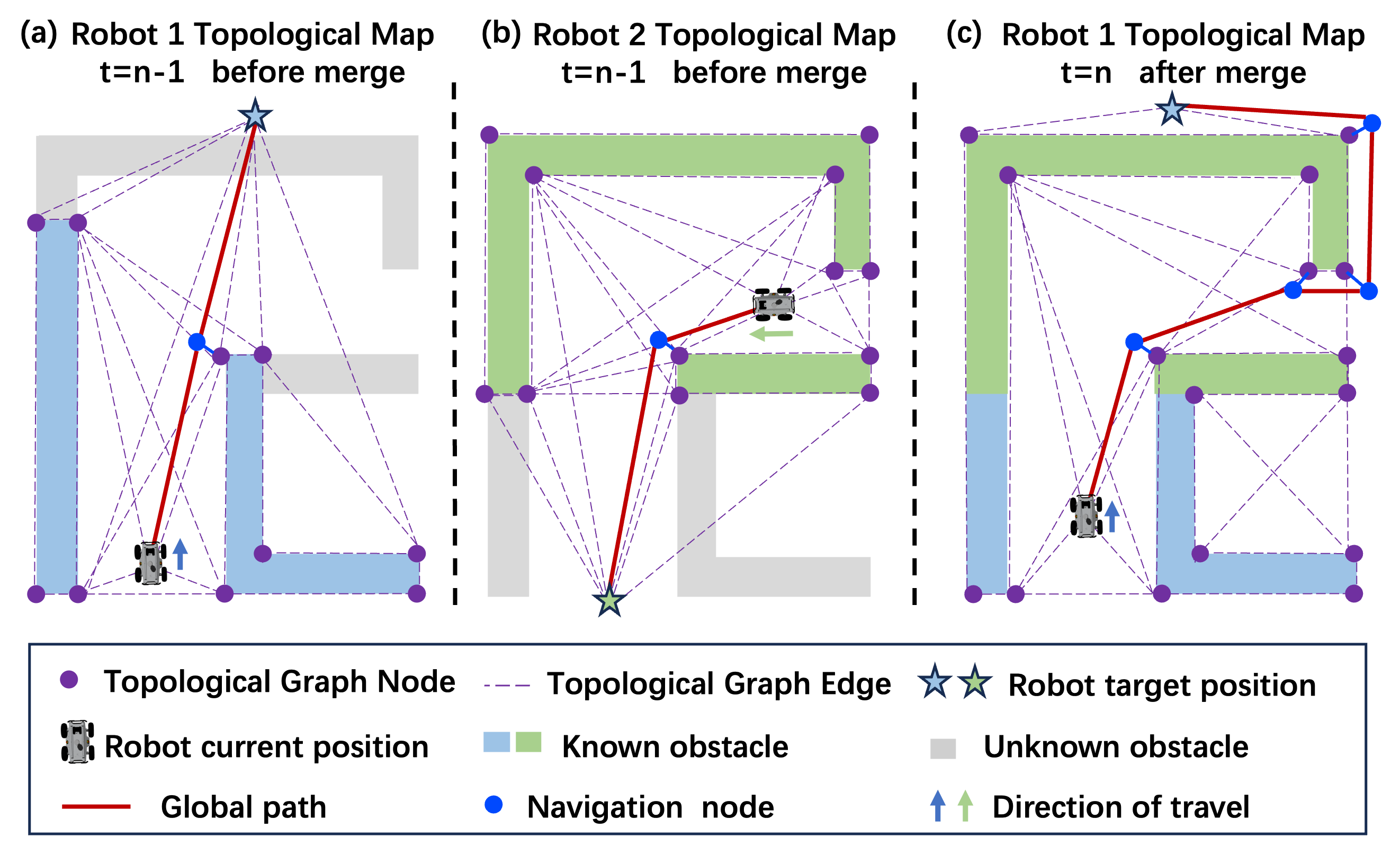}
    \caption{The topology sharing and guidance process. At time t-1, Robot 1 merges its own topology map (blue) with the received topology map (green) from Robot 2 and re-plans its global path based on the newly expanded knowledge.}
    \label{fig:topology sharing}
    \vspace{-1em}
\end{figure}
\subsection{Topology Sharing and Global Guidances (TSGG)}

To achieve robust global navigation without a shared metric map, the TSGG layer constructs and shares a lightweight topological representation of the environment. We model the topology as a visibility graph, as it compactly captures the connectivity of navigable space and is resilient to minor sensor noise. The process, outlined in Alg.~\ref{topology_construction} and depicted in Fig.~\ref{fig:topology sharing}, enables the team to build a collective environmental understanding, plan globally efficient paths that avoid large-scale traps, and provide strategic direction to the local planner.

Our hierarchical map representation is inspired by \cite{farplanner}, our primary contribution lies in the fully decentralized mechanism for pairwise map fusion and knowledge aggregation.
For each robot $i$, the environment's topology at time $t$ is represented by a graph $\mathcal{G}_{i,t} = (\mathcal{V}, \mathcal{E})$, where the vertex set $\mathcal{V}$ includes the vertices of observed obstacle contours, the robot's start position, and its goal. An edge $(v_a, v_b) \in \mathcal{E}$ exists if the line segment connecting vertices $v_a$ and $v_b$ is collision-free.

\begin{algorithm}[t]
\SetKwFunction{expand}{expandAABB}
\SetKwProg{Fn}{Function}{:}{}
\caption{Topology Sharing and construction}\label{topology_construction}
\SetKwInOut{Input}{input}\SetKwInOut{Output}{output}
\Input{Lidar point cloud $\mathcal{P}_{i,t}$ from robot $i$ at timestep $t$,
robot $i$'s topological map $\mathcal{G}_{global, t-1}^i$ and binary image $\mathcal{I}_{global, t-1}^i$ at timestep $t-1$,
received obstacle contours $\{\mathcal{C}_{other, t}^j\}_{j \neq i}$ from other robots $j$}
\Output{$\mathcal{G}_{global, t}^i$: Robot $i$'s updated topological map.}

$\mathcal{I}_{\text{ego}} \gets \text{ProjectToImage}(\mathcal{P}_{i,t})$ \tcp{Project local lidar scan to binary image}

$\mathcal{C}_{ego} \gets \text{FindContours}(\mathcal{I}_{\text{ego}})$ \tcp{Extract own obstacle contours}

$\mathcal{I}_{\text{fused}} \gets \text{ProjectToGridImage}(\mathcal{I}_{global, t-1}^i,\mathcal{C}_{ego})$ \tcp{Project local contours to binary image map}

\For{received obstacle contour $\mathcal{C}_{j, t}$ in $\{\mathcal{C}_{j, t}\}_{j \neq i}$}{
     $\mathcal{I}_{\text{fused}} \gets \text{FuseMaps}(\mathcal{I}_{\text{fused}}, \mathcal{C}_{j, t}$) \tcp{Fuse teammate contours into map}
    
}
 
$\mathcal{C}_{fused} \gets \text{FindContours}(\mathcal{I}_{\text{fused}})$ \tcp{Re-extract unified contours}
 
$\mathcal{G}^i_{\text{local,t}} \gets \text{BuildLocalGraphFromContours}(\mathcal{C}_{fused})$ \tcp{Build current local topology graph}
 
$\mathcal{G}_{global, t}^i \gets \text{MergeTopologies}(\mathcal{G}_{global, t-1}^i, \mathcal{G}^i_{\text{local,t}})$ \tcp{Merge local graph into the global graph}
\KwRet{$\mathcal{G}_{global, t}^i$}\;
\end{algorithm}

\subsubsection{Local Topological Map Construction}

The process of constructing a consistent local topological map begins with an robot $i$'s own perception. The Lidar point cloud, $\mathcal{P}_{i,t}$, is first projected into a local binary image, $\mathcal{I}_{\text{ego}}$, from which the robot's own obstacle contours, $\mathcal{C}_{\text{ego}}$ are extracted using the \texttt{FindContours} function based on \cite{opencv_findcontours}. Each closed-loop contour is an ordered sequence of points representing an obstacle boundary.
A key challenge in decentralized mapping is fusing partial observations from disparate, unaligned reference frames. To this end, we employ an image-space fusion technique. The robot projects both its own contours ($\mathcal{C}_{\text{ego}}$) and any received contours from teammates ($\{\mathcal{C}_{other, t}^j\}_{j \neq i}$) onto its existing global binary map, $\mathcal{I}_{global, t-1}^i$. This fusion process, performed by \texttt{ProjectToGridImage} and \texttt{FuseMaps}, accumulates all available spatial information into a single, consistent binary map, $\mathcal{I}_{\text{fused}}$.
From this map, a clean and consistent set of obstacle contours, $\mathcal{C}_{\text{fused}}$, is re-extracted. These contours serve as the foundation for a new, up-to-date local topological map, $\mathcal{G}^i_{\text{local,t}}$, generated by the \texttt{BuildLocalGraphFromContours} function.

\subsubsection{Global Topology Update}

Next, the newly constructed local graph $\mathcal{G}^i_{\text{local,t}}$, is integrated into the robot's persistent global map, $\mathcal{G}_{global, t-1}^i$. The \texttt{MergeTopologies} function performs this fusion by identifying and matching overlapping nodes and edges, seamlessly incorporating the new information into the existing global structure, following the method in \cite{farplanner}. This produces an updated, more comprehensive global map $\mathcal{G}_{global, t}^i$.

\subsubsection{Global Path Planning}

With the updated global map $\mathcal{G}_{global, t}^i$, the robot plans a strategic path from its current location to its goal using a depth-first search algorithm. The resulting path is a sequence of topological nodes that serves as a high-level reference, providing long-range guidance to the low-level CRTP layer.

\begin{figure}[t]
    \centering
    \includegraphics[scale=0.14]{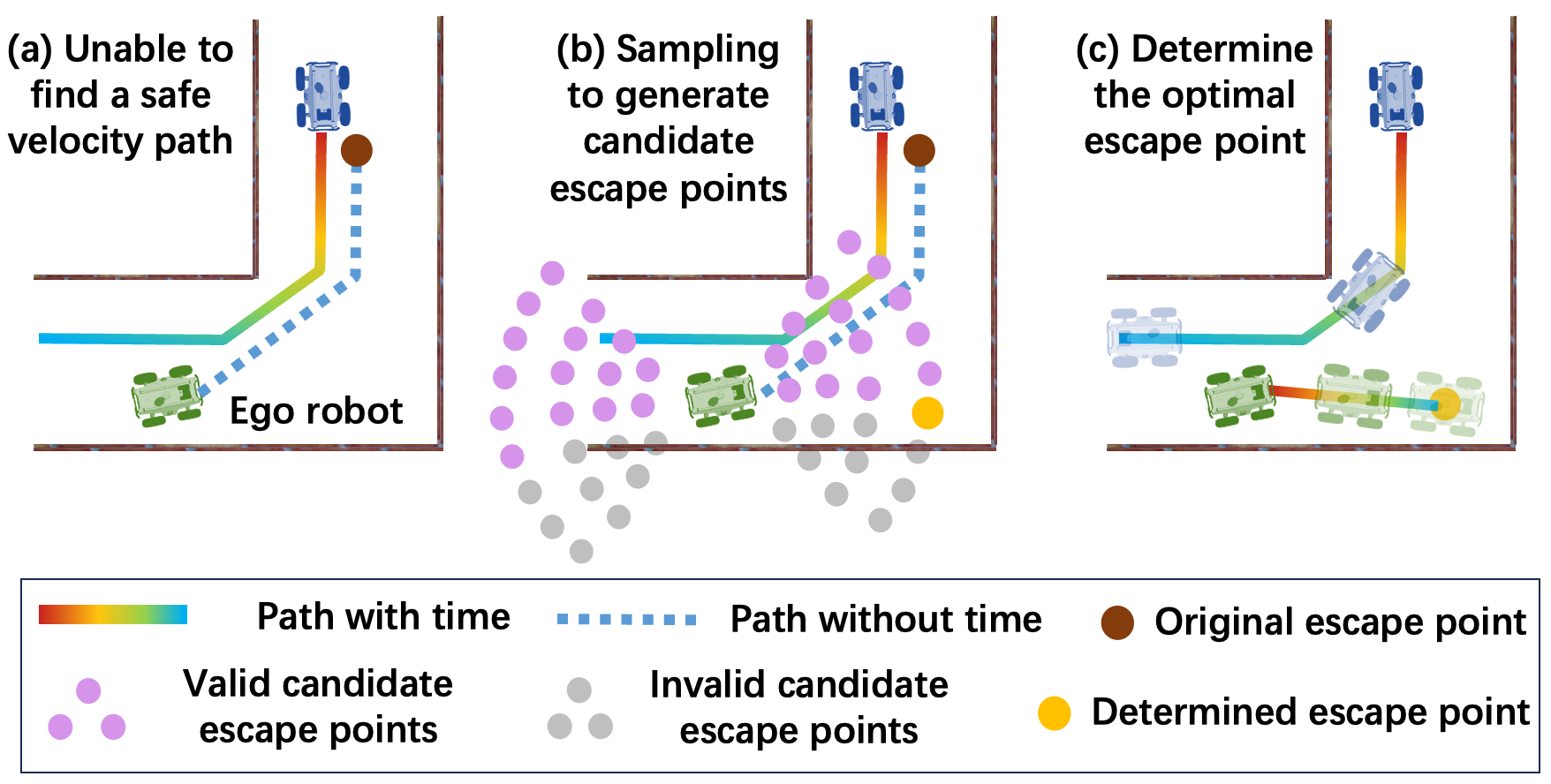}
    \caption{The local conflict resolution process. If the initial plan towards the local goal (red point) is blocked, the robot samples candidate escape points in forward and backward sectors. It evaluates these points based on a cost function and selects the optimal one (yellow point) to generate a new, conflict-free trajectory.}
    \label{conflict_resolution}
    \vspace{-1em}
\end{figure}

\subsection{Conflict Resolution and Trajectory Planning (CRTP)}

Inspired by the global path, there may be conflict situations between robots in some localized regions. The CRTP module translates the strategic guidance from the TSGG layer into a smooth, dynamically feasible, and collision-free trajectory. It operates in real-time to handle local interactions with static obstacles and other robots. The core of this module is a reactive planning pipeline that generates an initial path, uses a novel ‘escape point’ strategy to resolve deadlocks, and finally refines the path into an executable trajectory via nonlinear optimization.

\begin{figure*}[t]
    \centering
    \includegraphics[scale=0.20]{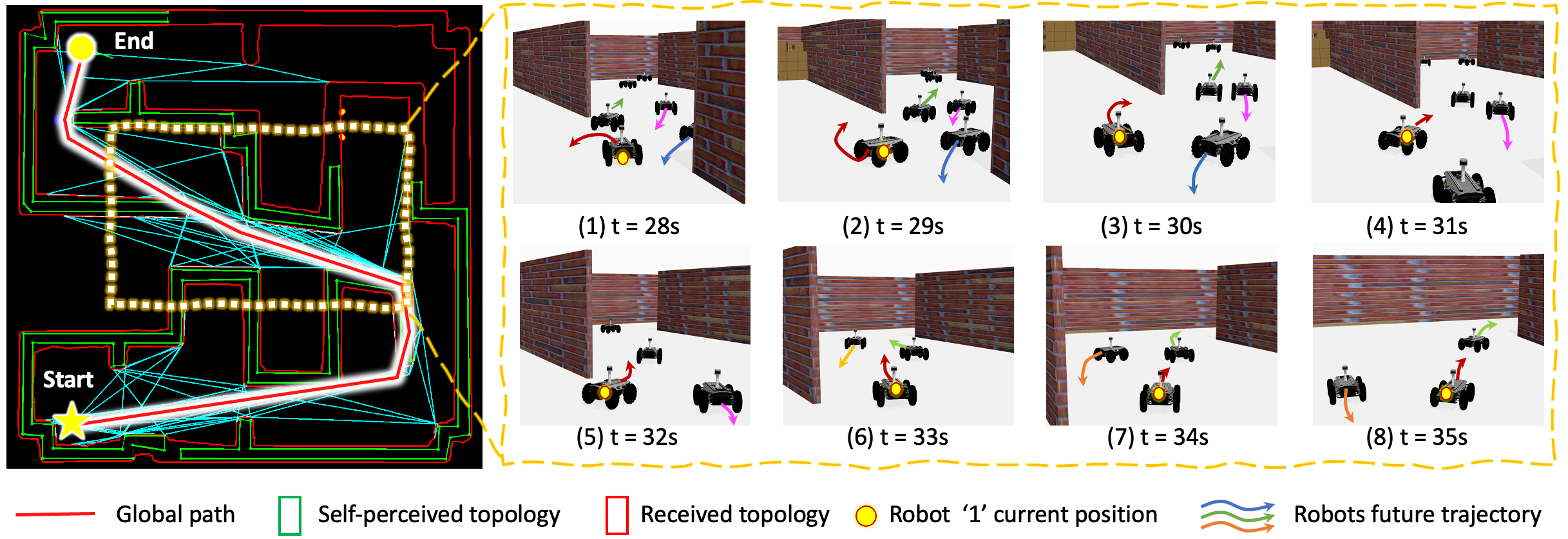}
    \caption{An example of the system in action. (Left) The topological map constructed by Robot 1 in Scenario A, showing a complex environment with multiple paths. (Right) A snapshot of Robot 1's first-person view as it navigates a dense central region, demonstrating effective local conflict avoidance.}
    \label{fig:robot1}
    \vspace{-1em}
\end{figure*}
\subsubsection{Conflict detection and resolution}
While following the global route, robots can enter dense configurations leading to deadlocks. Our CRTP module preemptively handles these situations. First, a local goal is selected from the global topological path based on a look-ahead distance. An initial path to this goal is planned using the A* algorithm, considering only static obstacles. 

This initial path is then checked for spatio-temporal conflicts with other robot' predicted trajectories using a Space-Time (ST) graph analysis \cite{Space-Time}. If the ST planner fails to find a valid velocity profile, a common occurrence in cluttered spaces where all time intervals are occupied, our escape point mechanism is triggered.
Instead of persisting towards the blocked goal, the system samples a set of candidate escape points to find a temporary diversion. As shown in Fig.~\ref{conflict_resolution}, these points are sampled uniformly in forward and backward sectors relative to the robot's heading. Each candidate point is evaluated via a cost function that balances progress towards the goal, safety, and path smoothness. The candidate with the minimum cost is selected as a temporary local objective, guiding the robot out of the deadlock until a feasible trajectory towards the original local goal can be found.

\subsubsection{Trajectory optimization and excution}
The initial conflict-free path from the previous stage provides a high-quality reference but may lack smoothness and dynamic feasibility. We refine it into an executable trajectory using optimization. The trajectory is represented as a piecewise polynomial, and we leverage the differential flatness of our robot model to formulate the task as an unconstrained nonlinear optimization problem.

The objective is to find the optimal polynomial coefficients $\boldsymbol{c}$ and total time $T$ that minimize a cost function balancing multiple criteria:
\begin{equation}
    \min_{\boldsymbol{c}, T} J = \int_{0}^{T} \lVert \boldsymbol{p}^{(3)}(t) \rVert^2 dt + w_T T + S_{\Sigma}(\boldsymbol{c}, T)
    \label{eq:objective}
\end{equation}
where $\boldsymbol{p}(t)$ is the robot's cartesian position curve, serving as the flat output. The first term penalizes the integral of the squared jerk, promoting trajectory smoothness. The second term, weighted by $w_T$, penalizes total time, encouraging efficiency. The final term, $S_{\Sigma}$, is a summation of penalties that enforce all system constraints as soft constraints, including:
\begin{itemize}
    \item \textbf{Static Obstacle Avoidance:} A penalty is imposed based on the distance to static obstacles observed in the local grid map. We use a differentiable distance function to ensure the robot maintains a safe margin from environmental hazards.
    \item \textbf{Kinodynamic Feasibility:} For our differential-drive robot, the linear velocity $v(t) = \lVert \dot{\boldsymbol{p}}(t) \rVert$ and angular velocity $\omega(t)$ are derived from the flat output and its derivatives. We impose penalties if they exceed the predefined limits $v_{\max}$ and $\omega_{\max}$.
    \item \textbf{Inter-Robot Collision Avoidance:} To ensure safety, we model each robot as a convex polygon and use a smooth, differentiable signed-distance function to penalize proximity between robots, ensuring collision-free paths.
\end{itemize}
This formulation allows the complex problem to be solved efficiently using gradient-based nonlinear solvers, and we adopt a fixed time-step discretization strategy to further improve computational performance in multi-robot scenarios.

\begin{figure*}[t]
    \centering
    \includegraphics[scale=0.28]{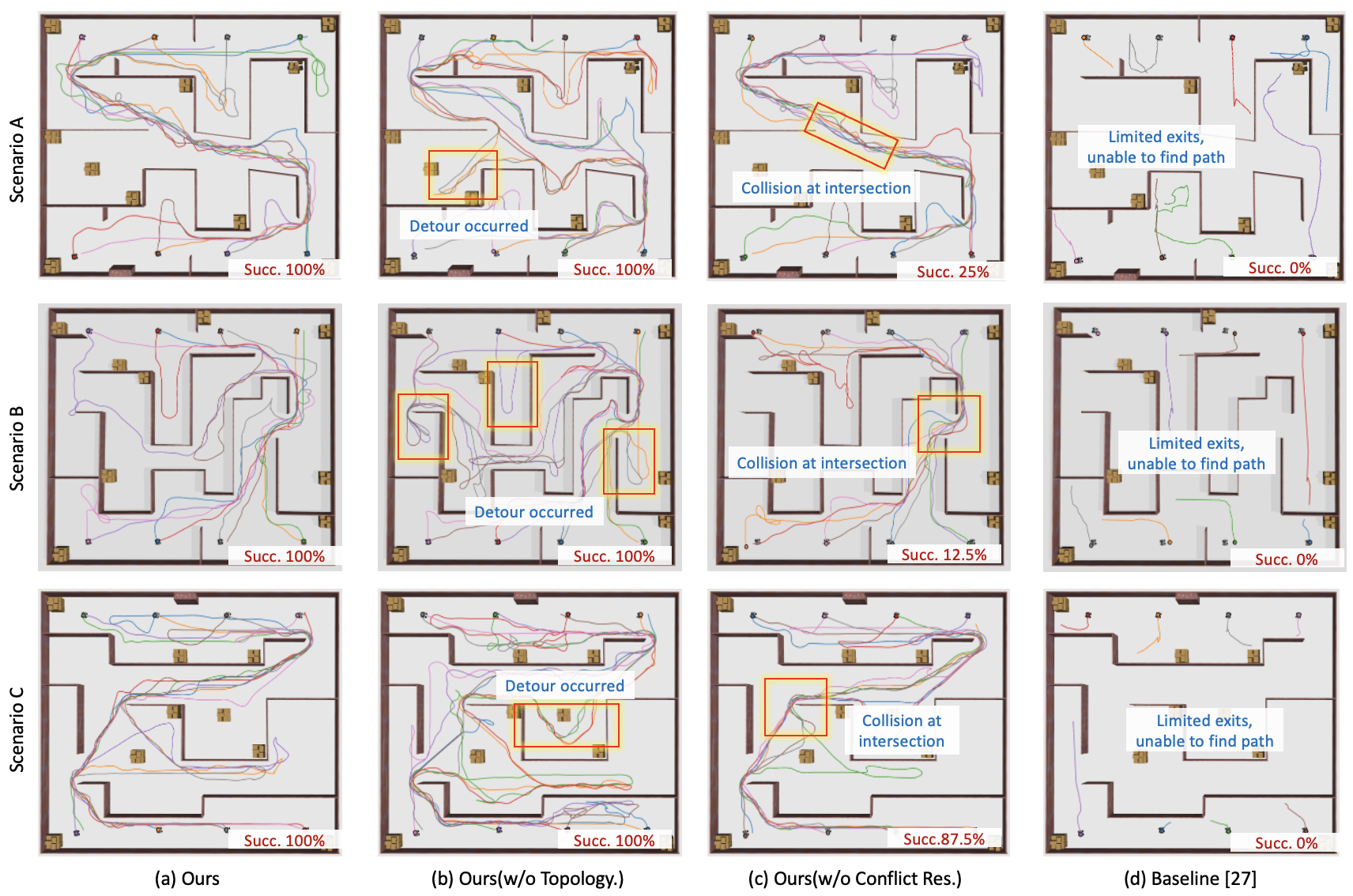}
    \caption{Qualitative results from three challenging simulation scenarios. The robots were assigned to opposite sides of the map and navigated toward each other in a crisscross pattern.  For each scenario, we show the final trajectories for: (a) Our full method, demonstrating smooth, efficient paths; (b) Our method without topology sharing, resulting in inefficient exploration of dead ends; (c) Our method without conflict resolution, leading to gridlock and collisions in central corridors; and (d) The baseline \cite{car-like}, showing robots cannot effectively navigate topologically complex environments.}
    \label{all_exp}
    \vspace{0.5em}
\end{figure*}

\begin{table*}[t]
\centering
\renewcommand{\arraystretch}{1.2}
\setlength{\tabcolsep}{3.5pt}
\caption{Quantitative Comparison of Simulation Results. Best performance is marked in \protect\colorbox{green}{pink}, second-best in \protect\colorbox{yellow}{yellow}.}

\label{main_results}
\begin{tabular}{@{}lcccc|cccc|cccc@{}}
\toprule
\multirow{2}{*}{\textbf{Method}} & \multicolumn{4}{c|}{\textbf{Scenario A}} & \multicolumn{4}{c|}{\textbf{Scenario B}} & \multicolumn{4}{c}{\textbf{Scenario C}} \\
 & Succ. (\%) & Time (s) & Len. (m) & Effi. & Succ. (\%) & Time (s) & Len. (m) & Effi. & 
 Succ. (\%) & Time (s) & Len. (m) & Effi. \\ \midrule
Ours (Full Method) & \cellcolor{green}100 & \cellcolor{yellow}47.67 & \cellcolor{yellow}74.14 & \cellcolor{yellow}0.51 & \cellcolor{green}100 & \cellcolor{yellow}64.48 & \cellcolor{green}99.8 & \cellcolor{green}0.34 & \cellcolor{green}100 & \cellcolor{yellow}69.89 & \cellcolor{yellow}111.91 & \cellcolor{yellow}0.31 \\
Baseline \cite{car-like} & 0 & - & - & - & 0 & - & - & - & 0 & - & - & - \\
Ours (w/o Topology) & \cellcolor{green}100 & 69.29 & 110.41 & 0.41 & \cellcolor{green}100 & 69.44 & 114.22 & 0.27 & \cellcolor{green}100 & 83.83 & 137.85 & 0.25 \\
Ours (w/o Conflict Res.) & \cellcolor{yellow}25 & \cellcolor{green}42.41 & \cellcolor{green}66.08 & \cellcolor{green}0.58 & \cellcolor{yellow}12.5 & \cellcolor{green}64.08 & \cellcolor{yellow}100.35 & \cellcolor{yellow}0.33 & \cellcolor{yellow}87.5 & \cellcolor{green}63.51 & \cellcolor{green}99.98 & \cellcolor{green}0.34 \\ \bottomrule
\end{tabular}
\end{table*}

\section{EXPERIMENTS}
To validate the effectiveness, robustness, and computational efficiency of our proposed decentralized navigation framework, we conducted a series of extensive experiments in both simulation and the real world. Our evaluation is designed to answer the following key questions:
\begin{itemize}
\item \textbf{Q1 (Overall Performance)}: Does our approach enable safe and efficient multi-robot navigation in unknown, structurally complex environments?
\item \textbf{Q2 (Value of Topology Sharing)}: Does decentralized, lightweight topology sharing measurably improve team navigation efficiency?
\item \textbf{Q3 (Value of Conflict Resolution)}: Can our proposed method effectively resolve local conflicts and deadlocks in congested areas?
\end{itemize}

\subsection{Experimental Setup}
Our simulation experiments were conducted in a custom 3D simulator built on Gazebo, allowing for precise control and repeatable scenarios. We designed multiple $40m \times 40m$ cluttered maps containing numerous obstacles, narrow corridors, and large dead-end structures to challenge the planners. The robot team consists of 8 differential-drive robots with a rectangular footprint of $0.4m \times 0.8m$, a maximum linear velocity of $2.0m/s$, and a maximum angular velocity of $\pi/2 rad/s$. Each robot is equipped with a simulated Lidar with a $360^{\circ}$ field of view and a 4m perception range. All algorithms were executed on a desktop computer with an Intel Core i9-14900 CPU and 16 GB of RAM.

We evaluate the performance of our framework using the following primarily quantitative metrics:
\begin{itemize}
    \item \textbf{Success Rate (\%):} The percentage of trials where all robots reach their designated goals without any collision (inter-robot or with static obstacles). A higher rate indicates greater safety and reliability.
    \item \textbf{Average Travel Time (s):} The average time taken for all robots to reach their goals in a successful trial, measuring mission timeliness.
    \item \textbf{Average Travel Distance (m):} The average distance traveled by all robots, reflecting planning efficiency.
    \item \textbf{Path Optimality:} The average ratio of the Euclidean start-to-goal distance to the actual path length for all robots. Values closer to 1 indicate more direct, optimal paths.
\end{itemize}

\subsection{Simulation Results}
\subsubsection{Quantitative Analysis in Challenging Scenarios} We designed three challenging simulation scenarios, each with 8 robots tasked with crossing from one side of the map to the other. This setup intentionally creates dense traffic and forces interactions in the map's central corridors. Furthermore, the environments contain large U-shaped dead-end regions, which act as topological traps. We compare our full framework against a state-of-the-art reactive baseline \cite{car-like} and two ablated versions of our own method. The comprehensive results are presented in Table~\ref{main_results} and Fig.~\ref{all_exp}. Additionally, Fig.~\ref{fig:robot1} illustrates the topology map generated by Robot 1 in Scenario A, along with local obstacle avoidance details from Robot 1's first-person perspective in the central region of the map.

Answering Q1, our full approach achieved higher success rates across three challenging scenarios. This demonstrates its ability to robustly synergize global strategic planning with local tactical maneuvering, ensuring no robot gets permanently trapped or deadlocked. The reactive baseline \cite{car-like} failed, with a 0\% success rate. Lacking mechanism for inter-robot communication or shared environmental understanding, the robots are fundamentally "short-sighted." They are unable to reason about the global topology, leading them to become irreversibly stuck in the dead-end traps, as shown in Fig.~\ref{all_exp}(b). This result indicates that reactive strategies are difficult to employ for solving navigation problems in environments with complex topological features.

\begin{figure}[t]
    \centering
    \includegraphics[scale=0.14]{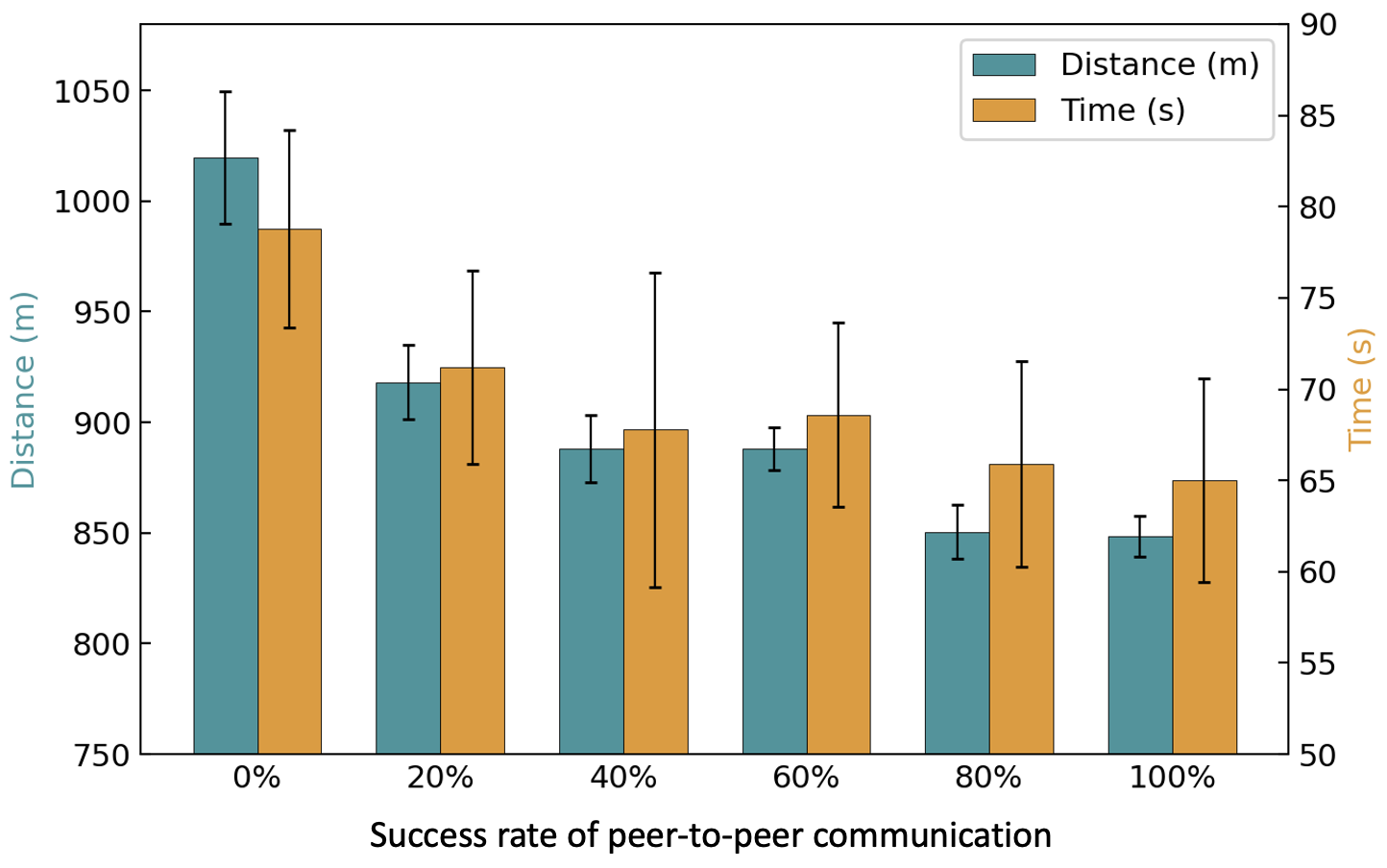}
    \caption{System performance under degraded communication. Path length and completion time increase gracefully as the communication success rate drops, demonstrating robustness to packet loss.}
    \label{fig:ablation}
\end{figure}

\subsubsection{Ablation Studies}
To isolate the contributions of our key modules, we conducted an ablation study, with results also detailed in Table~\ref{main_results} and Fig.~\ref{all_exp}.
\begin{itemize}
    \item \textbf{w/o Topology Sharing:} Each robot relies solely on its local sensor data for global pathfinding, without benefiting from the shared environmental knowledge.
    \item \textbf{w/o Conflict Resolution:} The dynamic escape-point mechanism is disabled; the planner relies only on basic velocity adjustments and trajectory optimization for avoidance.
\end{itemize}

To answer Q2, the w/o Topology Sharing variant, stripped of its global reasoning capability, achieved a 100\% success rate but with drastically reduced efficiency. Lacking global guidance, robots frequently enter and explore the large dead-end traps before backtracking (Fig.~\ref{all_exp}). This behavior is reflected in its metrics: compared to our full method, its average path length and completion time increased by up to 26.8\% and 23.2\%, respectively. This clearly demonstrates that the TSGG layer is critical for achieving high navigational efficiency.
Addressing Q3, the w/o Conflict Resolution variant failed catastrophically in dense scenarios, with its success rate plummeting to a mere 12.5\% in Scenario B due to frequent inter-robot collisions and deadlocks (Fig.~\ref{all_exp}). Although it sometimes recorded shorter travel times in its few successful runs (by avoiding the small detours our escape strategy introduces), its inability to reliably resolve conflicts makes it impractical. This outcome validates that our CRTP module is indispensable for ensuring safety and mission success in dense multi-robot scenarios.

\begin{table}[t]
\centering
\renewcommand{\arraystretch}{1.0}
\setlength{\tabcolsep}{6.5pt}
\caption{System communication and computational overhead.}
\label{overhead_results}
\begin{tabular}{@{}cccc@{}}
\toprule
\makecell{\textbf{Topology} \\ \textbf{BW.}} & 
\makecell{\textbf{Topology} \\ \textbf{fusion T.}} & 
\makecell{\textbf{Global} \\ \textbf{plan T.}} & 
\makecell{\textbf{Trajectory} \\ \textbf{opt T.}} \\
\midrule
6.29KB/s & 2$\sim$3ms & 1$\sim$2ms & 22$\sim$80ms \\
\bottomrule
\end{tabular}
\end{table}

\subsubsection{Robustness and Overhead Analysis}
We further analyzed the system's robustness to communication degradation by varying the peer-to-peer message success rate. As shown in Fig.~\ref{fig:ablation}, the framework demonstrates a certain degree of tolerance in the event of packet loss.  A drop in communication rate from 100\% to 20\% resulted in only a 12\% increase in completion time, as infrequent topological updates are still sufficient for effective strategic planning. Performance degrades sharply at 0\% communication rate, which is equivalent to the "w/o Topology Sharing" ablation.

Furthermore, we measured the system's overhead under ideal (100\%) communication conditions. As shown in Table~\ref{overhead_results}, this low overhead further demonstrates our method's tolerance in communication-constrained scenarios.

\subsection{Real-World Demonstration}
\begin{figure}[t]
    \centering
    \includegraphics[scale=0.15]{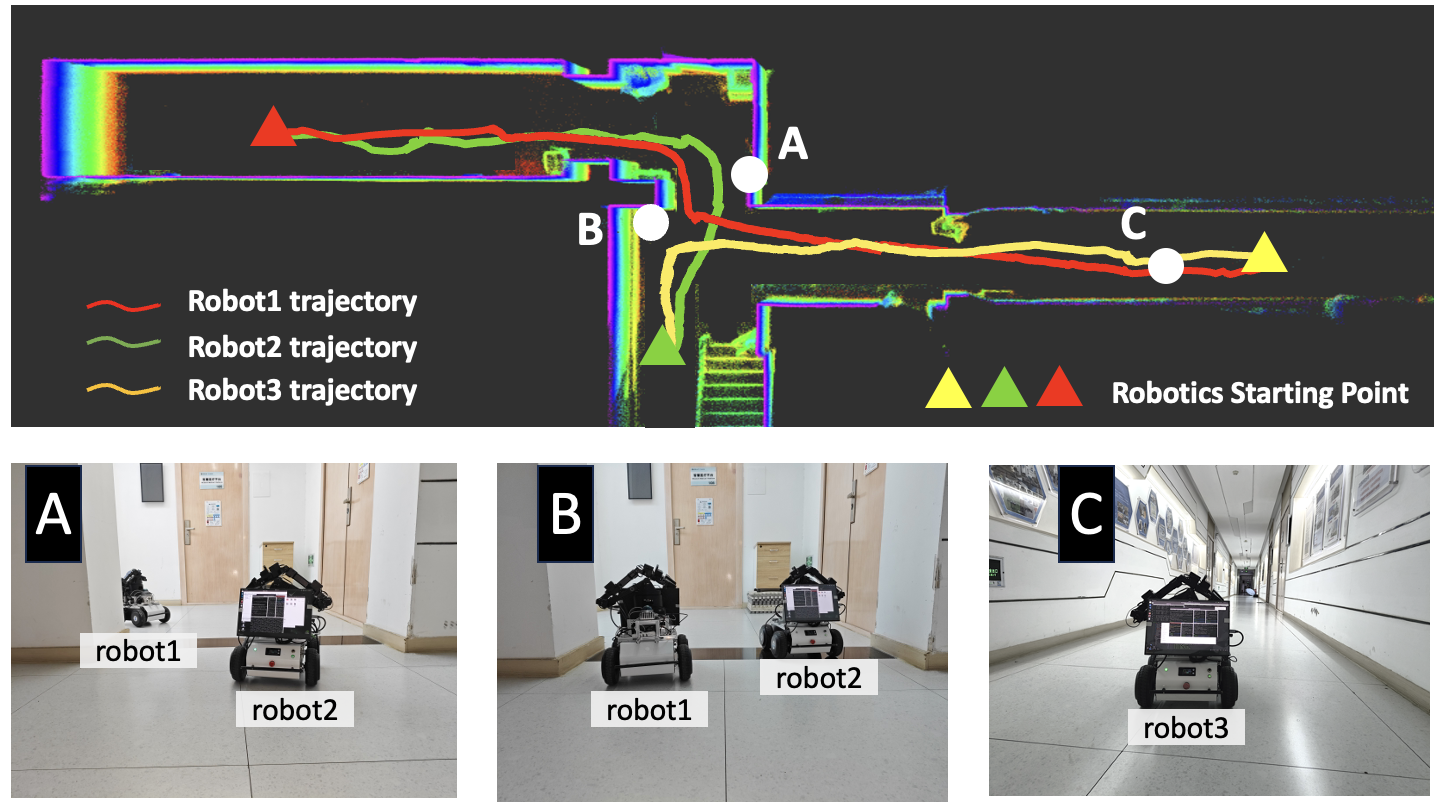}
    \caption{Real-world experiment with 3 robots in a corridor environment. The figure shows the robot's situation in areas A, B, and C.}
    \label{fig:real_world}
\end{figure}

For real-world validation, we deployed our system on a team of three robots. Each robot was equipped with an onboard Jetson Orin NX for all computations and a Mid-360 LiDAR for perception. Odometry was provided by the Fast-LIO SLAM algorithm \cite{fast-lio}. Relative localization between robots was achieved by matching sparse features in their LiDAR scans. The sparse point cloud features was insufficiently dense to fully capture environmental contours. Following pose estimation, robots exchanged their locally-constructed, lightweight topological map.  Communication between robots was achieved via the mesh self-organizing network communication module. As shown in Fig.~\ref{fig:real_world}, the robots were assigned a traversal scenario requiring them to navigate through a corridor and cross paths in the central area. The system demonstrated robust performance, with all robots safely reaching their targets during the trial. The physical demonstrations validated our framework's transferability from simulation to reality and its effectiveness in handling real-world sensor noise and communication delays. Experimental videos are available in the supplementary materials.

\section{CONCLUSIONS}
In this paper, we introduced a decentralized, hierarchical navigation framework that effectively synergizes long-range topological guidance with real-time metric-level planning. Through extensive simulations and real-world experiments, we have demonstrated its ability to overcome the limitations of purely reactive or centralized approaches, enabling robust, efficient, and deadlock-free navigation in complex, unknown environments under communication constraints. Our work shows that by decoupling the problem into strategic and tactical layers, a multi-robot team can achieve emergent global awareness and local agility without a shared global coordinate system.
Future work will focus on scaling our system to larger teams in more expansive and dynamic environments. We also plan to explore leveraging the shared topological map as a scaffold for richer semantic information sharing, enabling robots to collaborate on more complex tasks beyond navigation.

\vspace{1.0em}





\bibliographystyle{unsrt}
\bibliography{reference}

\end{document}